\documentclass[conference]{IEEEtran}
\IEEEoverridecommandlockouts
\usepackage{cite}
\usepackage{amsmath,amssymb,amsfonts}
\usepackage{algorithmic}
\usepackage{graphicx}
\usepackage{textcomp}
\usepackage{xcolor}
\usepackage{hyperref}
\usepackage{booktabs}
\usepackage{graphicx}
\usepackage{balance}

\def\BibTeX{{\rm B\kern-.05em{\sc i\kern-.025em b}\kern-.08em
    T\kern-.1667em\lower.7ex\hbox{E}\kern-.125emX}}
\begin{document}

\title{Do LLMs \textit{Understand} Ambiguity in Text? A Case Study in Open-world Question Answering} 


\author{\IEEEauthorblockN{Aryan Keluskar}
\IEEEauthorblockA{\textit{School of Computing \& AI} \\
\textit{Arizona State University} \\
Tempe, AZ, USA \\
akeluska@asu.edu}
\and
\IEEEauthorblockN{Amrita Bhattacharjee}
\IEEEauthorblockA{\textit{School of Computing \& AI} \\
\textit{Arizona State University} \\
Tempe, AZ, USA \\
abhatt43@asu.edu}
\and
\IEEEauthorblockN{Huan Liu}
\IEEEauthorblockA{\textit{School of Computing \& AI} \\
\textit{Arizona State University} \\
Tempe, AZ, USA \\
huanliu@asu.edu}
}


\maketitle

\begin{abstract}
Ambiguity in natural language poses significant challenges to Large Language Models (LLMs) used for open-domain question answering. LLMs often struggle with the inherent uncertainties of human communication, leading to misinterpretations, miscommunications, hallucinations, and biased responses. This significantly weakens their ability to be used for tasks like fact-checking, question answering, feature extraction, and sentiment analysis. Using open-domain question answering as a test case, we compare off-the-shelf and few-shot LLM performance, focusing on measuring the impact of explicit disambiguation strategies. 
We demonstrate how simple, training-free, token-level disambiguation methods may be effectively used to improve LLM performance for ambiguous question answering tasks. We empirically show our findings and discuss best practices and broader impacts regarding ambiguity in LLMs.
\end{abstract}


\section{Introduction}

Recent years have seen unprecedented advancements in the development of large language models (LLMs). Today, LLMs are ubiquitous and easily accessible for use by the general public - either via platforms that allow API calls, such as the OpenAI API\footnote{https://platform.openai.com}, or through openly available model weights for open LLMs, such as via Huggingface\footnote{https://huggingface.co/models}. Since late 2022, large and powerful LLMs have taken over the world of written communication with at least 56\% of students using AI in their college work according to a survey \cite{Nam_2023}. Most of these students, and people overall, harness the conversational capability of this AI for tasks such as problem solving and question-answering. Agentic AI workflows have also started to increase in popularity ~\cite{forbes1}, where these LLMs are used for NLP tasks such as sentiment analysis and data annotation.

However, human language is highly context-dependent and complex. Much of the meaning in language, both spoken and written, comes from the context in which it is used, as well as social and psychological cues. This makes it challenging for LLMs to grasp human language, which otherwise would be simple and straightforward for human listeners or readers to understand. 
LLMs often struggle with the inherent uncertainties of human communication, leading to misinterpretations, miscommunications, and biased responses which weaken their trust and ability to be used for real-world tasks.
\textit{Ambiguity} in natural language poses significant challenges to Large Language Models: much recent work has demonstrated how LLMs struggle to understand ambiguous text in prompts and instructions. This is particularly challenging when lay users prompt LLMs for solving tasks, obtaining answers to trivia questions, etc. - in such cases, the LLM may fail to `understand' the context and either fail to respond properly, or even hallucinate a factually wrong response with high confidence \cite{zhang2023sirenssongaiocean}. Given the importance of evaluating the sensitivity of LLM to ambiguity in context, in this work we use open-domain question answering as a test case to compare the off-the-shelf LLM performance on ambiguous questions. In our analyses, we further explore simple, training-free methods for disambiguating questions and compare such performance with naive prompting. Through experiments on two state-of-the-art LLMs with a publicly available ambiguous question-answering dataset, we present interesting insights and discuss implications and best practices.


\begin{figure}
    \centering
    \includegraphics[width=1\linewidth]{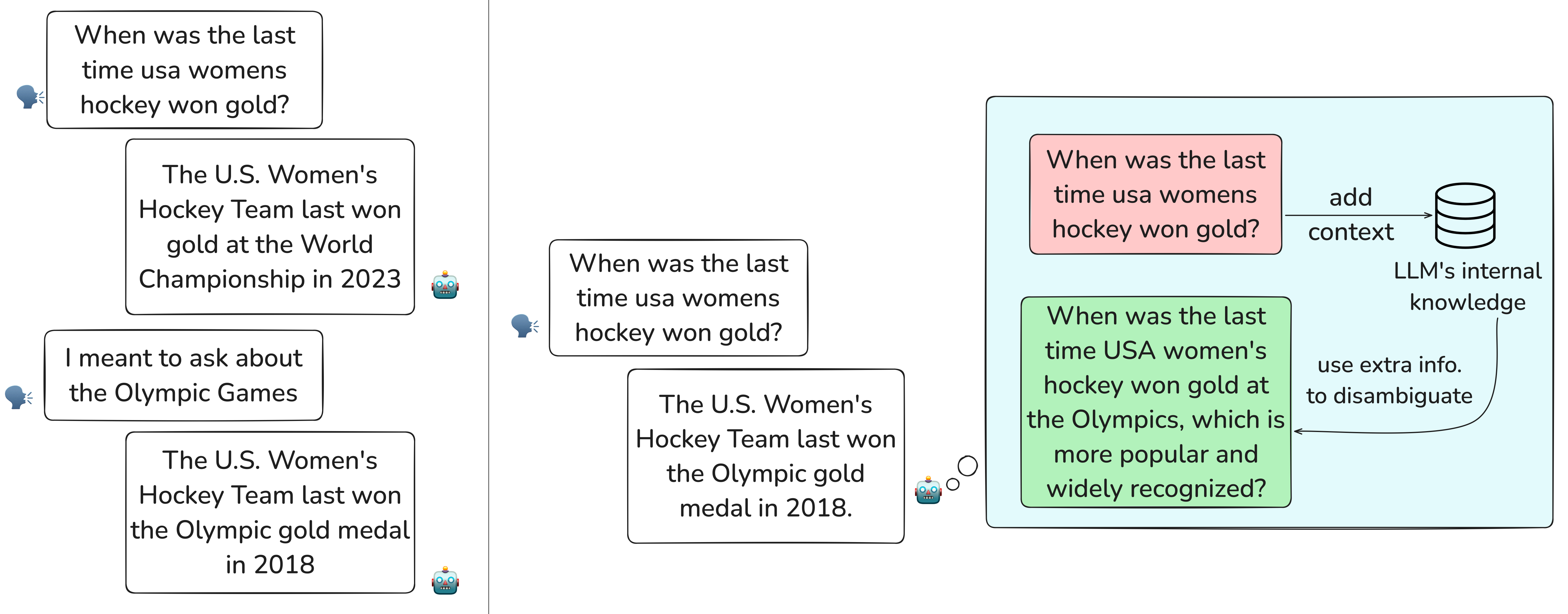}
    \caption{The problem of \textit{ambiguity} in open domain question answering (QA) (left), and how we try to solve it for large language model QA (right).}
    \label{fig:enter-label}
\end{figure}

\section{Background and Related Work}


Large language models are complex neural network based models\cite{vaswani2017attention} that are capable of generating human-like text~\cite{bubeck2023sparks}. Most recent LLMs consist of transformer-based architectures, with a huge number of parameters, on the order of billions. Recent LLMs that are most widely used (such as OpenAI's GPT family of models~\cite{radford2019language,brown2020language,achiam2023gpt}, Meta AI's Llama~\cite{touvron2023llama,touvron2023llama2}, etc.) are trained on massive amounts of textual data scraped from the internet, then further fine-tuned using instruction-style data. Such a step is called instruction-tuning~\cite{zhang2023instruction} via which the LLM learns how to follow instructions in user prompts. While recent LLMs have achieved admirable levels of fluency and performance on a variety of natural language understanding (NLU) as well as general tasks such as code generation, solving math problems and even reasoning~\cite{bubeck2023sparks,bhattacharjee2024fighting,liu2024mathbench}, performance often depends on the way in which these models are prompted. The sensitivity of LLMs to variation in prompt is an active area of research.

Many recent works have looked into the sensitivity of LLMs to minor variations in the prompt~\cite{salinas2024butterfly}, such as with respect to the format~\cite{sclar2024quantifying}, etc. Prompts may also have various types of ambiguity; some recent work has tried looking at the response of LLMs to the task ambiguity in the prompt~\cite{tamkin2022task}. In particular, the aspect of ambiguity in language has also been explored~\cite{tamkin2022task}. However, although the concept of ambiguity in natural language has been explored significantly from the perspective of computational linguistics ~\cite{amb1,levesque2012winograd}, the effect on ambiguous language in prompts on the performance of LLMs is still under-explored.  Therefore in this work, we investigate LLM sensitivity to ambiguity in prompts, especially in the task of open-domain question answering.

\section{Problem Definition}


Ambiguity is inherent to open-domain question answering, making it very difficult to prompt these LLMs such that we get a single, unambiguous answer. The most popular LLMs fail to address this in their system cards, even though this can lead to misinformation and bias when social context is ignored while answering an ambiguous question. For example, someone asking `What is the home stadium of the Cardinals?' has different answers depending on whether the question is about `Arizona Cardinals in football' or `St. Louis Cardinals in baseball'. According to OpenAI, generation of content through a human-like tone increases hallucinations, which makes assessing an LLM on ambiguous questions in human language more important \cite{OpenAI_2024}. Given the prevalence of such inherent ambiguity in questions from a user prompting an LLM, we measure the performance of the LLM as a question-answering agent on (i) ambiguous questions such as the example above, and compare with (ii) disambiguated version of the question. More formally, say we are given a dataset $\mathcal{D}^{amb}$ with triples $(q_i, q^d_i, \hat{a}_i), i\in(1,n)$ where $q_i$ is the ambiguous question, $q^d_i$ is the disambiguated version of that question, and $\hat{a}_i$ is the ground truth answer for this question. Given an LLM $M$ of choice, we aim to compare $M(q_i)$ and $M(q^d_i)$ for $i\in(1,n)$, i.e. compare performance across ambiguous and disambiguated questions.

\section{Methodology and Experimental Settings}

We conduct a series of controlled experiments involving the two LLMs on a dataset of ambiguous real-world questions. Our approach emphasizes the evaluation of LLM sensitivity by measuring the effect of linguistic and contextual modifications on its output accuracy to answer ambiguous questions. We employed three distinct prompting strategies to generate answers from the selected LLMs: (1) a naive (or baseline) direct question-answering prompt, (2) a rephrasing strategy that attempts to add linguistic perturbation to the ambiguous question, and (3) a contextual enrichment approach that uses the model’s internal knowledge to disambiguate the given question. The selection of these strategies are aimed to comprehensively measure the models' accuracy to comprehend ambiguity through different types of perturbations. We apply these three prompting strategies to the LLMs over a randomly sampled subset of 1,000 ambiguous questions from a larger publicly available dataset of ambiguous question-answers. We also incorporated variations in temperature parameters to observe their impact on the consistency and accuracy of the model output; we used two temperature settings: default (1) and low (0.2) on a scale of 0 to 2. The outputs are collected, stored, and analyzed using two metrics to draw comparisons between the models' out-of-the-box performance and disambiguation methods. The metrics are chosen such that both the meaning and the wording of the final output are assessed. We draw comparative insights into the models' baseline performance and examine the effect of disambiguation techniques into mitigating the inherent challenges posed by ambiguous human language. We further perform a small-scale fine-tuning to evaluate whether task-specific disambiguation fine-tuning helps to improve performance.

\subsection{Dataset(s)}

Since this analysis is in open-domain question answering, we use the publicly available NQ-Open \cite{47761} Dataset by Google, which contains real-world queries issued to the Google search engine before January 2018. It has more than 300,000 question-and-answer pairs, with the answer manually annotated by referencing information from Wikipedia. Around 50\% of NQ-Open questions do not have an answer label because they are perceived as ambiguous, since the annotators found diverse sources of ambiguity such as event and entity references. This research focuses primarily on a subset of the dataset: AmbigQA \cite{min2020ambigqa} which covers 14,042 questions with a diverse set of ambiguities.

\textbf{AmbigQA} is divided into 3 subsets: \textit{dev}, \textit{train } and \textit{test}. The \textit{train} set contains 10,036 question-answer pairs, while \textit{dev} contains 2,002. Each element in \textit{test} and \textit{dev} includes a question, one or more answers, the original NQ answer, contents of the visited Wikipedia pages, search queries used to obtain evidence articles \& search results. However, each element in \textit{test} contains only a question and contents of the evidence articles in html and plaintext. NQ-Open and some subsets of AmbigQA also contain evidence documents attached to each question, which have been proven to be crucial to identify and characterize ambiguities. Each question in AmbigQA also has multiple human-provided disambiguated questions, where each one of them has a different context. For this research project, we have used a sample of 1,000 random questions from the train subset. The average length of a question in this sample is 8.93 words with 47.09 characters, and that of an answer is 2.30 words with 12.94 characters. Within this sample, 879 questions have one factual answer, 70 have two, 20 have three, and the rest have more than three answers. In AmbigQA, every plausible answer is paired with a disambiguated rewrite of the original question, such as specifically asking \textit{``Where is the home stadium of Arizona Cardinals in the NFL?"} This is particularly helpful in evaluating the limits of accuracy when used with question disambiguating techniques. 

\subsection{LLMs used}

We use two variants of state-of-the-art LLMs from OpenAI: \textbf{GPT-4o} \footnote{https://platform.openai.com/docs/models/gpt-4o} is a multimodal GPT model developed by OpenAI for real-time applications, offering enhanced capabilities to answer factual questions. \textbf{GPT-4o mini} \footnote{https://openai.com/index/gpt-4o-mini-advancing-cost-efficient-intelligence/} is a smaller and a cost-effective version of the GPT-4o model. This model has a context window of 128K tokens and has a knowledge cutoff of October 2023. Both of these models have been pre-trained on diverse datasets, including web text, scientific articles, and conversational data, to enhance their contextual understanding and generation capabilities. We use the gpt-4o and gpt-4o-mini model endpoints respectively on OpenAI API \footnote{https://platform.openai.com/docs/guides/text-generation} to access these models with version as of October 25th. Additional parameters included top\_p set to 1 and presence\_penalty set to 0, which are both defaults in the API, and max\_completion\_tokens was left unspecified. Each experiment is executed using API calls within a few days of each other to minimize the impact of model updates.

\subsection{Disambiguation Methods} \label{disambiguation}

In our experiments, we used two prompt-level disambiguation methods. We experimented with a variety of prompts, and finally selected two appropriate prompts for disambiguating a given question using the two GPT models. We also analyze the effect of lowering the temperature parameter on the model's accuracy in answering ambiguous questions.

\textbf{Naive:} For each question, we prompt the out-of-the-box LLM to answer it as concisely as possible to get a baseline for our experiment. 

\vspace{1.5mm}
\noindent\fbox{%
    \parbox{0.95\columnwidth}{%
Answer the question as concisely as possible with ONLY one answer without any other text: 
\\
Question: \{$x_p$\} 
}}
\vspace{2mm}

\textbf{Rephrase using \textit{What}:} In preliminary experiments, we found that rephrasing a question to begin with ``what" makes it more specific than the initial ambiguous question, reducing the variability of responses. After obtaining the rephrased question, we pass it back to the LLM to obtain the final answer. For our random sample of 1,000 questions in these experiments, the average length of the rephrased question was 13.01 words, which is about 1.45 times longer than the original question.

\vspace{1.5mm}
\noindent\fbox{%
    \parbox{0.95\columnwidth}{%
Rewrite this question replacing all questions with a what, but retain the meaning by specifying what entity or what person or what timeframe the ``what" is answering. Also, specify the current year is 2018 if needed to answer a time-based question. 
\\
Question: \{$x_p$\} 
}}
\vspace{2mm}


\textbf{Adding Context to the Ambiguous Question:} Since LLMs have vast amounts of world knowledge due to the extensive pre-training and instruction tuning done on them, we use that world knowledge from LLMs to find and return relevant information about the ambiguous question. This approach was chosen to reduce the effect of scope ambiguity \cite{kamath2024scope}, which can be caused by differing semantic structures. We are able to differentiate between two sentences with the same structure but different meanings because of the general knowledge we have about the entities it is referring to. After generating context about the question, we append the original ambiguous question at the end of this information blob and pass it back to the LLM to obtain the final answer. For our random sample of 1,000 questions in these experiments, the average length of the question with added context was 126.90 words, which is approximately 14.2 times longer than the original question.

\vspace{1.5mm}
\noindent\fbox{%
    \parbox{0.95\columnwidth}{%
Add extra information to the following question. Also specify the current month and year is January 2018, so answer questions accordingly. Your aim is to disambiguate what it is asking.
\\
Question: \{$x_p$\} 
}}
\vspace{2mm}







\subsection{Evaluation Metrics}

We measure the effect of our disambiguation methods on the overall accuracy of the LLMs by using semantic similarity between the LLM responses and the ground truth responses. We do this over measuring token overlap directly since several instances have ground truth answers that may be rephrased in multiple ways, all of which are correct. This allows for more meaningful evaluations by taking phrasing variations into account. Specifically, we use OpenAI's \textit{text-embedding-3-large}\footnote{https://platform.openai.com/docs/guides/embeddings} vector embedding model to generate the vectors and then computed the cosine similarity metric between two given texts. The distances are normalized on a scale from 0 to 1, where a value of 1 indicates that the compared texts are identical, while lower values suggest varying degrees of divergence. In our experiments and results, we measure:

1. \textbf{Distance Between Ambiguous and Disambiguated Questions:} This metric quantifies the semantic shift introduced by each disambiguating prompt in the question. Since we have an upper bound for the performance of an LLM with disambiguation, this metric also allows us to observe how effective were our prompts in changing the question in comparison to the given disambiguated question.

2. \textbf{Distance Between Baseline Answer and Disambiguated Answer:} This measures the difference between the responses generated by the LLM when given the original ambiguous question and the disambiguated question. A larger distance in this case implies that the model's answer remains relatively consistent across different contexts and phrasings, indicating robustness while answering ambiguity.

3. \textbf{Distance Between Baseline Answer and Ground Truth:} This metric acts as a reference point to evaluate the initial effectiveness of the out-of-the-box LLM answers before any disambiguation. This provides insights into how closely the model approximates human-annotated answers without additional prompts. Comparing this distance with distances involving disambiguated strategies is how we measure improvements in accuracy achieved through the proposed prompting techniques.

4. \textbf{Distance Between Ground Truth and Disambiguated Answer:} This metric measures the effectiveness of each disambiguation method in bridging the gap between model-generated answers and human-annotated ground truth answers. An increase in this metric after disambiguation indicates that the chosen prompts and contextual enrichment techniques have been able to pivot the LLM towards generating more accurate answers to ambiguous questions.


\section{Results and Discussion}

\begin{figure}
    \centering
    \includegraphics[width=1\linewidth]{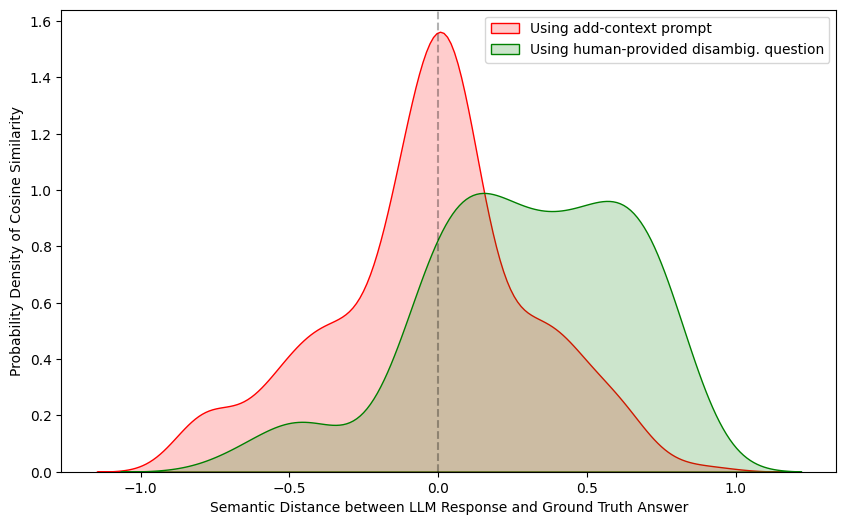}
    \caption{Kernel Density Estimate (KDE) Plot to compare the Cosine Similarity between Ground Truth Answer and LLM Response for the 2 disambiguation strategies for the randomly sampled subset of 1,000 Ambiguous Questions.}
    \label{fig:whole-context}
\end{figure}

\begin{figure}
    \centering
    \includegraphics[width=1\linewidth]{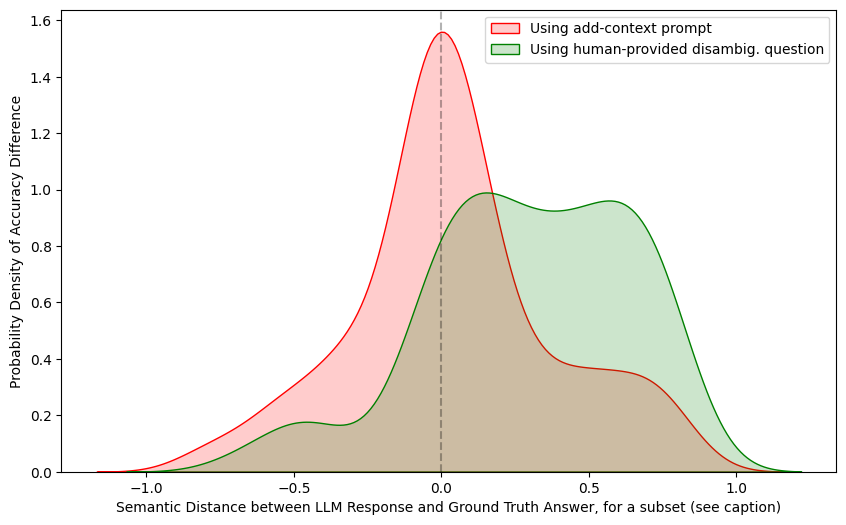}
    \caption{Kernel Density Estimate (KDE) Plot to compare the Cosine Similarity between Ground Truth Answer and LLM Response for the 2 disambiguation strategies for the subset of AmbigQA where human-provided answer for human-provided disambiguated question matched the ground truth. }
    \label{fig:subset-context}
\end{figure}


\begin{table}[]
\centering
\caption{Performance of GPT-4o on ambiguous and disambiguated questions. }
\label{tab:main-4o}
\resizebox{\columnwidth}{!}{%
\begin{tabular}{@{}ccccc@{}}
\toprule
\textbf{Metric}                                                     & \textbf{Naive} & \textbf{\begin{tabular}[c]{@{}c@{}}Disamb.\\  via `what'\end{tabular}} & \textbf{\begin{tabular}[c]{@{}c@{}}Disamb. \\ via context\end{tabular}} & \textbf{\begin{tabular}[c]{@{}c@{}}Upper-bound (via GT\\ disamb. questions)\end{tabular}} \\ \midrule
\begin{tabular}[c]{@{}c@{}}Question \\ coherence\end{tabular}       & -              & 0.905                                                                  & \multicolumn{1}{c|}{0.743}                                              & 0.317                                                                   \\ \midrule
\begin{tabular}[c]{@{}c@{}}Naive Answer \\ Overlap \end{tabular} & -              & 0.826                                                                  & \multicolumn{1}{c|}{0.799}                                              & 0.895                                                                   \\ \midrule
\begin{tabular}[c]{@{}c@{}}GT Answer\\  Overlap $\uparrow$\end{tabular}    & 0.759          & 0.778                                                                  & \multicolumn{1}{c|}{\textbf{0.789}}                                              & 0.858                                                                   \\ \bottomrule
\end{tabular}%
}
\end{table}

\begin{table}[]
\centering
\caption{Performance of GPT-4o-mini on ambiguous and disambiguated questions.}
\label{tab:main-mini}
\resizebox{\columnwidth}{!}{%
\begin{tabular}{@{}ccccc@{}}
\toprule
\textbf{Metric}                                                     & \textbf{Naive} & \textbf{\begin{tabular}[c]{@{}c@{}}Disamb. \\ via `what'\end{tabular}} & \textbf{\begin{tabular}[c]{@{}c@{}}Disamb. \\ via context\end{tabular}} & \textbf{\begin{tabular}[c]{@{}c@{}}Upper-bound (via GT\\ disamb. questions)\end{tabular}} \\ \midrule
\begin{tabular}[c]{@{}c@{}}Question \\ coherence\end{tabular}       & -              & 0.907                                                                  & \multicolumn{1}{c|}{0.739}                                              & 0.317                                                                   \\ \midrule
\begin{tabular}[c]{@{}c@{}}Naive Answer \\ Overlap \end{tabular} & -              & 0.771                                                                  & \multicolumn{1}{c|}{0.739}                                              & 0.745                                                                   \\ \midrule
\begin{tabular}[c]{@{}c@{}}GT Answer \\ Overlap $\uparrow$\end{tabular}    & 0.692          & 0.707                                                                  & \multicolumn{1}{c|}{\textbf{0.71}}                                               & 0.783                                                                   \\ \bottomrule
\end{tabular}%
}
\end{table}

In order to understand how leading LLMs tackle ambiguous questions in a question answering task, we investigate and aim to answer the following research question: 

\begin{itemize}
    \item \textbf{RQ1}: How well do off-the-shelf LLMs perform on ambiguous QA in a zero shot setting and does training free-disambiguation help?
    \item \textbf{RQ2}: Does few-shot fine-tuning improve performance?
    \item \textbf{RQ3}: Does reducing the temperature for LLM generation help in improving performance?
\end{itemize}

\textbf{RQ1}: We show the results for RQ1 for GPT-4o and GPT-4o-mini in Tables \ref{tab:main-4o} and \ref{tab:main-mini} respectively. In both tables, \textbf{Question coherence} refers to the semantic similarity between the ground truth disambiguated question and the ambiguous question when disambiguated via the LLM following one of the two methods; \textbf{Naive Answer Overlap} refers to the semantic similarity between LLM responses obtained via the disambiguating prompts vs. the naive prompt; and finally \textbf{GT Answer Overlap} refers to the semantic similarity between the LLM response and the ground truth answer in the dataset. Ideally, we want higher values for this metric. Interestingly, we see that for both GPT 4o and 4o-mini, using simple disambiguating prompts improves performance over the naive setting, implying that simple prompt-based, training-free approaches may be useful in improving LLM performance for ambiguous queries. Out of the two simple disambiguating methods explored, we see that disambiguation via adding context performs better for both LLMs.

\textbf{RQ2}: To evaluate whether small scale fine-tuning helps in improving LLM performance on ambiguous questions, we perform few-shot fine-tuning on GPT 4o-mini \footnote{owing to the lower cost of fine-tuning the \textit{mini} version of GPT-4o.}. To adapt our model for handling ambiguous questions, we fine-tuned the model using OpenAI's API. We randomly sampled 50 question-answer pairs from AmbigQA. Each ambiguous question was stored as the prompt for the LLM, with ground truth being stored as the expected response from the LLM. The file was formatted as shown below for the 50 questions. Using this, we initiated a fine-tuning job on OpenAI's fine-tuning API \footnote{https://platform.openai.com/docs/guides/fine-tuning/what-models-can-be-fine-tuned} which returned a model checkpoint. We used this fine-tuned model ID instead of gpt-4o-mini within our baseline prompt configuration.

\vspace{1.5mm}
\noindent\fbox{%
    \parbox{0.95\columnwidth}{%
\{``messages": \\  \{``role": ``user", ``content": \textless ambiguous question\textgreater\}, \\ \{``role": ``assistant", ``content": \textless ground truth answer\textgreater "\}\}
}}
\vspace{2mm}

To evaluate the performance of this fine-tuned model, we sample 1,000 ambiguous questions from the dataset at random and compare the performance between naive prompting on the 4o-mini model and naive prompting on the fine-tuned 4o-mini model. The \textbf{GT Answer Overlap} for the 4o-mini model is 0.643 while that for the fine-tuned 4o-mini model is 0.626. Therefore, we see that fine-tuning, at least at this small scale, does not provide any improvement in LLM performance on ambiguous questions. This reinforces our insight that simple training-free prompting methods for disambiguation work well in improving performance.

\begin{figure}
    \centering
    \includegraphics[width=0.99\columnwidth]{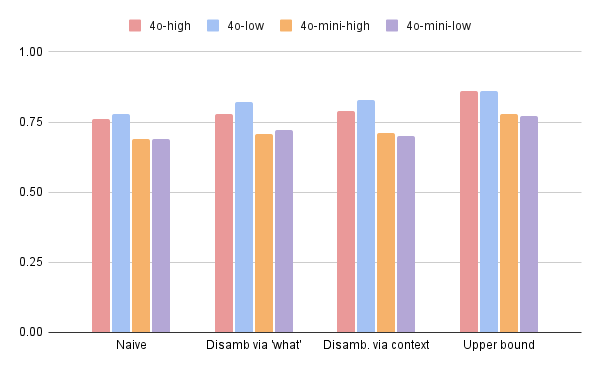}
    \caption{Comparison of \textbf{GT Answer Overlap} for GPT 4o and 4o-mini for both high and low temperatures. High = 1.0, low = 0.2. Higher overlap scores are better.}
    \label{fig:temp}
\end{figure}

\textbf{RQ3}: Using a lower temperature value for LLM generation results in reducing the `stochasticity' of the generated text, whereby the variance is reduced and the generated text is more predictable over multiple runs. As an ablation, we evaluate whether lower values of temperature help in improving performance of the LLM on ambiguous questions. We show the results for this in Figure \ref{fig:temp}: we see that although lower temperature (0.2 instead of 1.0, in this case) seem to have minor improvements in some cases, the difference is not that significant. This implies simply using a lower value of temperature may not provide any benefits in LLM performance for answering ambiguous questions.

\subsubsection*{Problem with naive contextual enrichment} The Figures \ref{fig:whole-context} and \ref{fig:subset-context} show why the average is not going up when an LLM is prompted to insert context into a question. Although adding context should skew the plot \ref{fig:whole-context} to the right (ie: be more similar to the ground truth), but instead its skewed to the left since its being held back every time it adds the wrong context which is not a problem when we have simply rephrased the question. Then, we take a subset of AmbigQA where the human-provided answer of a human-provided disambiguated question exactly matches the ground truth answer. Here, we see that plot \ref{fig:subset-context} of contextual enrichment does skew to the right. This shows that LLMs are able to better understand certain social cues to correctly disambiguate the provided question in cases where the human annotator was able to disambiguate them as well.

\section{Conclusion and Future Works} \label{future}


Our results indicate that contextual enrichment has the ability to significantly enhance model disambiguation accuracy, but it is often inaccurate because it tends to add irrelevant context to questions, making them impossible to fix by prompting. However, when we took a subset of AmbigQA where the human-provided answer of a human-provided disambiguated question provided matches the ground truth, adding context to those questions increases the accuracy of the model. Therefore, our analysis shows that even though LLMs struggle with ambiguity in prompts, simple training-free prompt-based disambiguation methods may help significantly in improving the performance of the LLM. 

In future work, we plan to fine-tune the LLM for accurate context-enhancement. We will take the contextually enriched information blob and fine-tune the model to generate a disambiguated question that is as close as possible to human-provided disambiguation to maximize accuracy for question-disambiguation based strategies. This will help in increasing the accuracy overall since LLMs in the open-world do not have access to disambiguated versions of a question, so through this fine-tuning we expect the LLM to learn social cues while disambiguating a question, which will improve the accuracy for questions outside of AmbigQA dataset.
We also plan to assess these prompt-based disambiguation techniques in open-source models such as Llama-3.1-8B-Instruct and Mixtral-8x7B~\cite{jiang2023mistral}, as well as to test the performance of our fine-tuned model on general factual question-answering by using SimpleQA by OpenAI and other questions from the NQ-Open dataset. This is a promising way forward to reduce hallucinations caused by ambiguity and better guard for an LLM's prompt sensitivity when its tasked with answering ambiguous questions.

\balance
\section*{Limitations}

This study adopted a general form of ``ambiguity" in analyzing the performance of Large Language Models. We recognize that a more thorough investigation of various types of ambiguity—such as multiple answers, time-dependent interpretations, multiple answer types, and other specific forms—could provide deeper insights into how a model falters in different kinds of ambiguity. 

Additionally, we suspect that our fine-tuning approach may have underperformed due to catastrophic forgetting \cite{luo2024empiricalstudycatastrophicforgetting}. As suggested in Section \ref{future}, future work could adopt more targeted and intentful fine-tuning strategies, such as development of a dedicated question disambiguation model or the application of linguistic refinements that current LLMs cannot perform in a zero-shot setting. Future studies could also mitigate the effects of catastrophic forgetting with methods such as Singular Value Decomposition \cite{franke2024preserving}. Implementing such strategies could enhance the stability and effectiveness of fine-tuned models in answering ambiguous questions.



\bibliographystyle{IEEEtran.bst}
\bibliography{main}

\end{document}